\documentclass{article}

\usepackage{amsmath}
\usepackage{booktabs}
\usepackage{graphicx}
\usepackage[dblblindworkshop, final]{neurips_2026}

\workshoptitle{World Models in Physical AI}

\usepackage[utf8]{inputenc} 
\usepackage[T1]{fontenc}    
\usepackage{hyperref}       
\usepackage{url}            
\usepackage{booktabs}       
\usepackage{amsfonts}       
\usepackage{nicefrac}       
\usepackage{microtype}      
\usepackage{xcolor}         

\title{Interpreting Physics in Video Encoders using Multi-Object Interaction Videos}

\author{%
  Ayushi Mehrotra \\
  Caltech \\
  \texttt{amehrotra@caltech.edu} \\
  \And
  Sai Vidyaranya Nuthalapati \\
  \texttt{vidyaranya.ns@gmail.com} \\
}

\begin{document}

\maketitle

\begin{abstract}
Video encoder models are increasingly used as world models for physical AI, where a policy plans by rolling out the latent, so what physics the latent holds bounds what the agent can plan over.
Moreover, existing interpretability work probes quantities belonging to a single object; the
quantities that make physics relational---momentum transfer, restitution, friction---are
defined \emph{between} bodies and have not been examined. We probe relational physics attributes in two frozen
self-supervised video encoders (V-JEPA2 and VideoMAE) under two-body settings and test if they transfer across environments. Our early findings are mixed. Both encoders separate
elastic from inelastic collisions reliably. V-JEPA2's own predictor is less likely to predict a
collision that creates kinetic energy compared by one that loses it. Momentum-violation
detection appears in only one of the two encoders, a kinetic-energy readout survives a
change of environment for V-JEPA2 but not for VideoMAE, and neither model holds a friction
coefficient that survives a change of geometry. 
\end{abstract}

\section{Introduction}
\label{sec:intro}

Video encoders are increasingly used as world models for physical AI, where latent representations support prediction, planning, and control \citep{garrido2025intuitive, assran2025vjepa2}. An agent that plans by rolling out such a model is constrained by the physical information its representations capture. Therefore, understanding which physical quantities are encoded is essential to understanding what the model can support at planning time.

Most of the evidence on this point is behavioral. Benchmarks score whether a model
predicts, answers, or prefers the physically correct continuation
\citep{riochet2018intphys, yi2020clevrer, bear2021physion, upadhyay2026worldbench}, and
recent encoders do well on them. However, a score cannot say which internal quantity
produced it. A representation that encodes a physical quantity and one that only tracks a
visible correlate of that quantity will perform identically on any evaluation drawn from
the environment the model was trained on. The two come apart only when the correlate stops
predicting the quantity, which is what a change of scene, geometry, or embodiment does at
deployment. Out-of-distribution evaluation does not settle this either, since any
environment we can construct carries its own correlates. Hence, deciding which of
the two a video encoder has learned requires evidence from the representation itself.

A newer line of work does open the models up, probing physical variables layer by layer
\citep{joseph2026interpreting, punzo2026layerwise, esmati2026invisible, alam2026steering}.
These studies primarily investigate properties of individual objects, such as position, speed, direction, and acceleration, or whether a scene is physically plausible. However, many physical interactions are governed by relational quantities. For instance, momentum transfer and restitution describe interactions between two bodies, while friction characterizes interactions between a body and a surface. These quantities are important for manipulation and driving, where successful planning requires reasoning about contact and its physical consequences. Unlike individual-object attributes, they cannot be characterized by considering either object in isolation. To our knowledge, prior work has not directly measured whether such relational quantities are encoded in video representations.

We investigate relational physics in two frozen self-supervised video encoders, VideoMAE-base \citep{tong2022videomae} and V-JEPA2-ViT-L \citep{assran2025vjepa2}, on
two-object collisions rendered in Kubric \citep{greff2022kubric}.We examine whether their representations capture physical properties of these interactions, including whether collisions are elastic or inelastic, whether collisions conserve momentum, how kinetic energy changes during a collision, and how friction affects motion. The models use different pretraining objectives: pixel reconstruction for VideoMAE and latent-space prediction for V-JEPA2. We compare representation-based readouts against a matched observer with access to visible kinematic quantities, and test whether readouts transfer unchanged across different scene families.

Our findings are mixed. Both encoders reliably distinguish elastic from inelastic collisions (AUCs of $0.911$ and $0.973$), but this distinction can be recovered from a ratio of visible speeds that does not account for mass, and therefore does not establish that the models represent kinetic energy. For kinetic energy, fitted probes alone do not establish whether the models distinguish physically impossible energy creation from other collision outcomes. We therefore examine the models’ own prediction and reconstruction mechanisms without training an additional readout. V-JEPA2’s predictor is more surprised by collisions that create kinetic energy than by those that lose it, whereas VideoMAE’s decoder shows the same trend without statistical significance. A kinetic-energy readout trained on sliding cubes transfers to colliding spheres for V-JEPA2 but performs at chance for VideoMAE. For momentum, violation detection is evident in only one encoder. Finally, neither encoder yields a friction-coefficient readout that transfers across environments with different floor angles.

\section{Experimental Setup}
\label{sec:prelim}
\vspace{-5pt}
\paragraph{Models.}
We study two frozen self-supervised video encoders: \textbf{VideoMAE-base} \citep{tong2022videomae} and \textbf{V-JEPA2-ViT-L}
\citep{assran2025vjepa2}.
\vspace{-8pt}
\paragraph{Datasets and pooling.}
All videos are rendered in Kubric \citep{greff2022kubric} with the simulator's exact
physical state recorded per frame, so every probe target is a ground-truth quantity. Because
our questions are about quantities defined \emph{between} objects, activations are pooled
over nine regions: three spatial windows---the whole frame (\texttt{all}) and each object
(\texttt{A}, \texttt{B}), taken from the renderer's instance masks projected onto the token
grid---crossed with three temporal windows---before contact (\texttt{pre}), after it
(\texttt{post}), and the whole clip (\texttt{all}). We write these \texttt{all.pre},
\texttt{A.post}, and so on. Both encoders attend bidirectionally over the full clip, so a
``pre'' token has already attended to the aftermath and no cell isolates prediction; we
treat the temporal split as a pooling choice rather than a causal one.

\paragraph{Probing.}
Probes are ridge regressions from a single pooled activation vector to a physical
target, scored out-of-fold. Continuous targets are scored by out-of-fold $R^2$ and binary
targets by AUC. Every reported probe is refit under
$12$ fold seeds and we give the mean and standard deviation across them.

\paragraph{Controls.}
Two controls accompany every result. An \textbf{oracle} reads the visible scalars (like velocities of both objects) which an observer can see without a notion of mass or energy, fitted through the identical
pipeline and folds. An fitted probe is informative only insofar as it beats the oracle. A \textbf{permutation null} refits the whole pipeline on
permuted labels and takes the maximum over the searched grid of layers and pools,
correcting reported $p$-values for that search.

We organize the experiments around two types of physical interactions. Section \ref{sec:collisions} examines two-body collisions, first describing the dataset construction and then evaluating whether the encoders capture three physical properties: momentum conservation, elasticity, and kinetic energy. Section \ref{sec:friction} follows the same structure for friction, introducing a separate cube–floor dataset before evaluating whether friction coefficients are recoverable from the representations and whether the corresponding readouts generalize across floor geometries.

\section{Two-Body Collisions}
\label{sec:collisions}

We use two-body collisions as a testbed for studying whether video encoders capture physical properties of multi-object interactions, focusing on momentum conservation, elasticity, and kinetic energy. In Appendix~\ref{sec:single},we examine the individual-object attributes underlying these quantities such as velocity, speed, direction, radius, and mass. Firstly, we find that they are highly decodable from object-specific tokens. Secondly, whether these attributes appear to emerge with network depth depends on how the representations are pooled, rather than on the encoder itself. Finally, frame-wide pooling preserves information about the sum of the two objects’ attributes but not their difference.

\subsection{Dataset Construction}
\label{sec:dataset}
All clips show two spheres of independently sampled radius and density colliding on a frictionless plane. We render the scenes in Kubric and record the simulator’s exact physical state for every frame. We construct the collision outcomes by editing the spheres’ velocity changes at contact, allowing us to vary momentum conservation and restitution independently while keeping the pre-contact trajectories identical between matched clips.

For a frictionless collision, the sum and ratio of the magnitudes of the two velocity changes satisfy
\[
|\Delta v_A| + |\Delta v_B| = (1+e)\,u_n, \qquad
\frac{|\Delta v_B|}{|\Delta v_A|} = \frac{m_A}{m_B},
\]
where \(v_A\) and \(v_B\) are the velocities of spheres \(A\) and \(B\) immediately before contact, \(n\) is the unit vector along the contact normal, and \(u_n = |(v_A-v_B)\cdot n|\) is their relative approach speed along that normal. The coefficient of restitution \(e\) characterizes the ratio of the relative normal separation speed to the relative normal approach speed, while \(m_A\) and \(m_B\) denote the sphere masses. The quantities \(\Delta v_A\) and \(\Delta v_B\) are the respective velocity changes caused by the collision. The first equation determines the sum of the magnitudes of the velocity changes, while the second determines their relative magnitudes through the mass ratio. These are separable degrees of freedom, allowing either to be edited while preserving the other. To construct momentum-violating collisions, we preserve the sum while changing how the velocity-change magnitudes are distributed between the spheres. This preserves restitution and the antiparallel contact geometry exactly. Each violating clip is paired with a momentum-conserving clip that is bit-identical up to the contact frame, so nothing before contact distinguishes them. We construct three collision regimes: elastic collisions, inelastic collisions, and collisions that create kinetic energy while conserving momentum (see Appendix~\ref{sec:samples_dataset} for examples). 

\paragraph{Elastic and inelastic collisions.}
Two cells of $75$ momentum-conserving clips and their $75$ violating twins, drawn over the
same sampling distribution and differing only in restitution: $e = 1.0$, where kinetic
energy is conserved exactly, and $e = 0.15$, where it is lost to deformation and the
post- to pre-collision ratio spans $0.098$--$0.992$ depending on the incoming geometry.

\paragraph{Collisions that create energy.}
Kinetic energy is not conserved in general---an inelastic collision legally converts it to
heat---so the physically impossible event is energy \emph{created}, which no simulator will
produce. We reach it with the complement of the edit above: rescaling both deflections by a
common factor $\lambda = (1 + e_{\mathrm{target}})/(1 + e_{\mathrm{sim}})$ holds the split
and moves the sum. Because $\lambda$ multiplies both objects,
$m_A \Delta v_A' + m_B \Delta v_B' = \lambda \cdot 0 = 0$ identically, so momentum is
conserved while restitution moves to its target. We create collision clips where $e = k$ with $k = 1.5$, giving kinetic-energy ratios of $1.043$--$2.029$, which violates physics.

\subsection{Momentum}
\begin{figure}[t]
\centering
\includegraphics[width=0.60\linewidth]{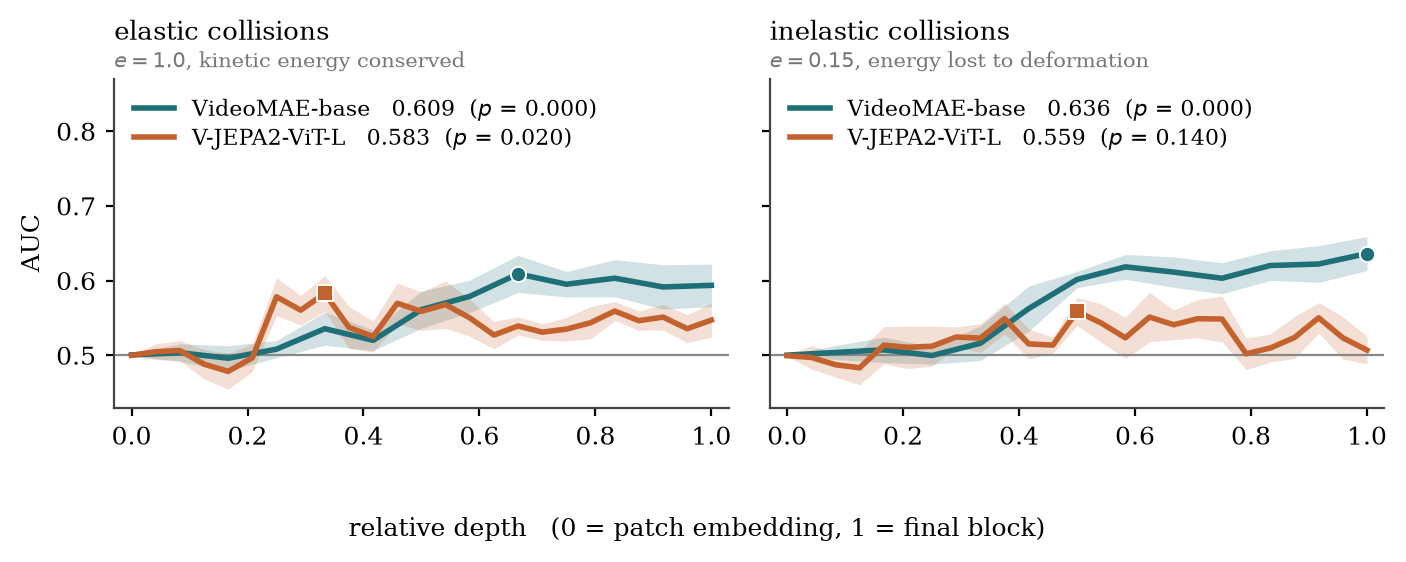}
\caption{\textbf{Momentum-violation detection across depth, by elasticity.} Out-of-fold AUC
of the norm route---regress the $\Delta p$ vector, take $|\cdot|$ outside the probem, pooled \texttt{all.pre}. The grey line is chance. Legend values give the
best layer with its $p$ corrected by a grid-max permutation null over all layers. VideoMAE clears the corrected null in both regimes; V-JEPA2 clears it marginally in
the elastic regime and not at all in the inelastic one. }
\label{fig:momentum}
\end{figure}

\paragraph{A two-stage readout.}
We test whether the encoders' representations contain information about momentum-conservation violations by predicting the net momentum change,
\begin{equation}
    \Delta p = m_A \Delta v_A + m_B \Delta v_B,
\end{equation}
where $m_A$ and $m_B$ are the masses of spheres $A$ and $B$, and $\Delta v_A$ and $\Delta v_B$ are their respective velocity changes during the collision. In our two-dimensional setup, $\Delta p \in \mathbb{R}^2$. Momentum is conserved when $\Delta p = (0,0)$; otherwise, its magnitude $\|\Delta p\|$ measures the violation.

The two classes have a geometric structure that makes direct linear classification difficult: momentum-conserving examples lie at the origin, while violating examples lie around it. Even a linear classifier given the exact momentum-change vector achieves an AUC of only 0.573, whereas a classifier given its exact norm achieves an AUC of 1.000. The distinction therefore depends on taking the vector norm, a nonlinear operation. We use a two-stage readout: first, ridge regression predicts $\Delta p$ from the representation; then, we compute the norm of the predicted vector. Directly probing the binary violation label yields near-chance performance for both VideoMAE-base (AUC $=0.475 \pm 0.015$) and V-JEPA2-ViT-L (AUC $=0.504 \pm 0.010$).

\paragraph{Detection across depth.}
The norm route recovers the violation in VideoMAE and, at best marginally, in V-JEPA2
(Fig.~\ref{fig:momentum}). VideoMAE clears its corrected null in both regimes, reaching AUC
$0.609$ on elastic collisions and $0.636$ on inelastic ones against grid-max nulls of
$0.550$ and $0.557$ ($p = 0.000$), with $7$ and $8$ of its $13$ layers individually above
the null. V-JEPA2 reaches $0.583$ on elastic collisions against a null of $0.571$
($p = 0.016$ under a $500$-draw permutation null) and $0.559$ on inelastic ones against
$0.571$ ($p = 0.126$). We read that as marginal rather than positive: the margin is $0.011$,
only $2$ of $25$ layers clear the null, and the effect does not survive the change of
regime. Elasticity makes no difference to VideoMAE, and both curves sit at chance through the
first third of the network before rising.

\paragraph{Where the momentum change is represented.}
Appendix~\ref{sec:single} shows that object-local pools decode individual-object attributes more accurately than frame-wide pools. However, the net momentum change itself is poorly recovered from object-local representations: predicting $\Delta p$ from concatenated object-local pools gives $R^2 = -0.404$, compared with $R^2 = 0.426$ from frame-wide pooling. Object-local variants of the two-stage readout also remain near chance. Since $\Delta p = m_A \Delta v_A + m_B \Delta v_B$ is a \emph{sum}
over the two objects, and a frame-wide pool superposes them rather than discarding either, the interaction-level quantity is more accessible from representations that combine information across the scene than from representations that isolate the two objects.

\subsection{Elasticity}
\label{sec:elasticity-full}

\begin{figure}[ht]
\centering
\includegraphics[width=0.40\linewidth]{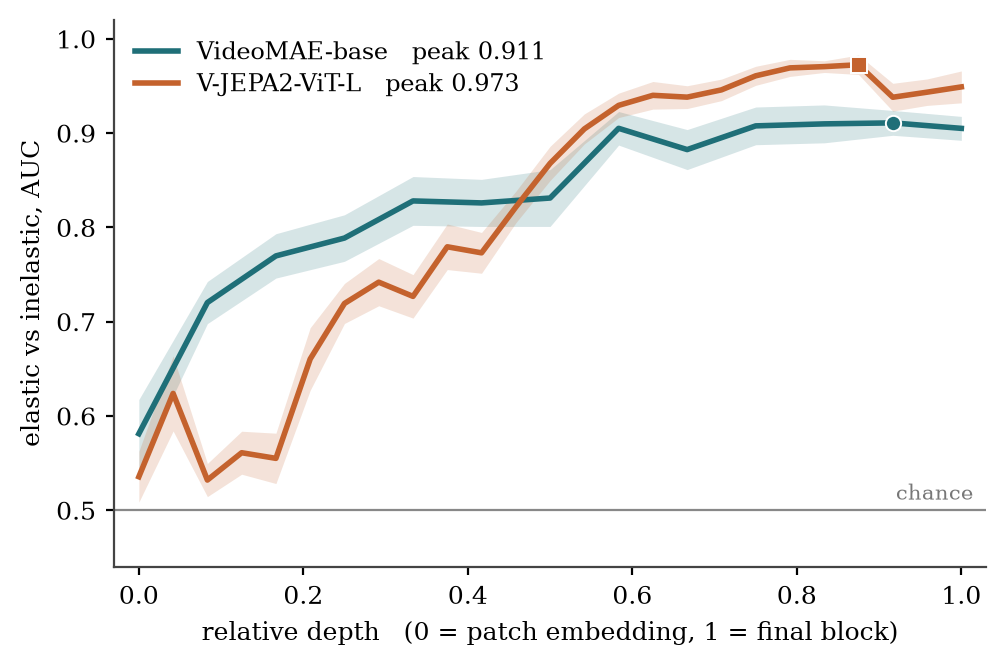}
\caption{\textbf{Elastic versus inelastic collisions across depth.} AUC from the
\texttt{all.post} pool on the $150$ momentum-conserving clips ($75$ per class), mean over
$12$ fold seeds with $\pm1$ sd bands. The distinction is built through the network, rising
from $0.581$ and $0.536$ at the patch embedding. Both peaks lie below an observer given the
contact geometry, which reads the same label at AUC $1.000$ from restitution alone.}
\label{fig:elasticity}
\end{figure}

\paragraph{Detection across depth.}
We first test whether the encoders can distinguish elastic from inelastic collisions, defining the binary label as $e < 1$, where $e$ is the coefficient of restitution. Both encoders separate the two classes reliably (Fig.~2): VideoMAE-base reaches an AUC of $0.911 \pm 0.013$ at block 10 of 12, while V-JEPA2-ViT-L reaches $0.973 \pm 0.010$ at block 20 of 24. Both improve substantially over their patch-embedding representations, which achieve AUCs of 0.581 and 0.536, respectively. Frame-wide representations outperform object-local representations throughout, consistent with restitution being a relational property of the collision pair.

\paragraph{Against a kinematic observer.}
High classification accuracy does not, by itself, establish that an encoder represents the physical quantities underlying the label. In our setup, the coefficient of restitution relates the relative separation speed to the relative approach speed along the contact normal:
\[
e =
\frac{|(w_A-w_B)\cdot n|}
     {|(v_A-v_B)\cdot n|},
\]
where $v_A$ and $v_B$ are the pre-contact velocities, $w_A$ and $w_B$ are the post-contact velocities, and $n$ is the unit contact normal. Because this ratio is mass-independent, the elastic-versus-inelastic label can be recovered from visible kinematics without explicitly representing kinetic energy.

To assess how much of the classification task can be explained by visible motion, we compare the encoders against kinematic observers with progressively richer inputs, fitted using the same pipeline and folds. A linear oracle using raw pre-collision velocity components achieves an AUC of $0.526 \pm 0.022$. Adding the four speeds raises the AUC to $0.733 \pm 0.015$, while a quadratic model of the velocity components reaches $0.887 \pm 0.020$. Both encoders outperform these baselines. However, an observer given velocities projected onto the contact normal achieves an AUC of 1.000, since restitution is determined by the ratio of the relative normal separation speed to the relative normal approach speed.

The contact normal itself is highly decodable from object-local representations, with $R^2=0.927$ for VideoMAE-base at block 8 using pool B.all and $R^2=0.943$ for V-JEPA2-ViT-L at block 15 using pool A.all. Together, these results show that the distinction between elastic and inelastic collisions is highly linearly decodable from both encoders, while an observer given velocities projected onto the contact normal can recover the label perfectly. However, successful classification does not establish that the encoders explicitly represent the coefficient of restitution; the observed performance may instead reflect other decodable features correlated with the elasticity label.

\subsection{Kinetic Energy}
\label{sec:energy}

\begin{figure}[t]
\centering
\begin{minipage}[b]{0.40\linewidth}
  \centering
  \includegraphics[height=3.3cm]{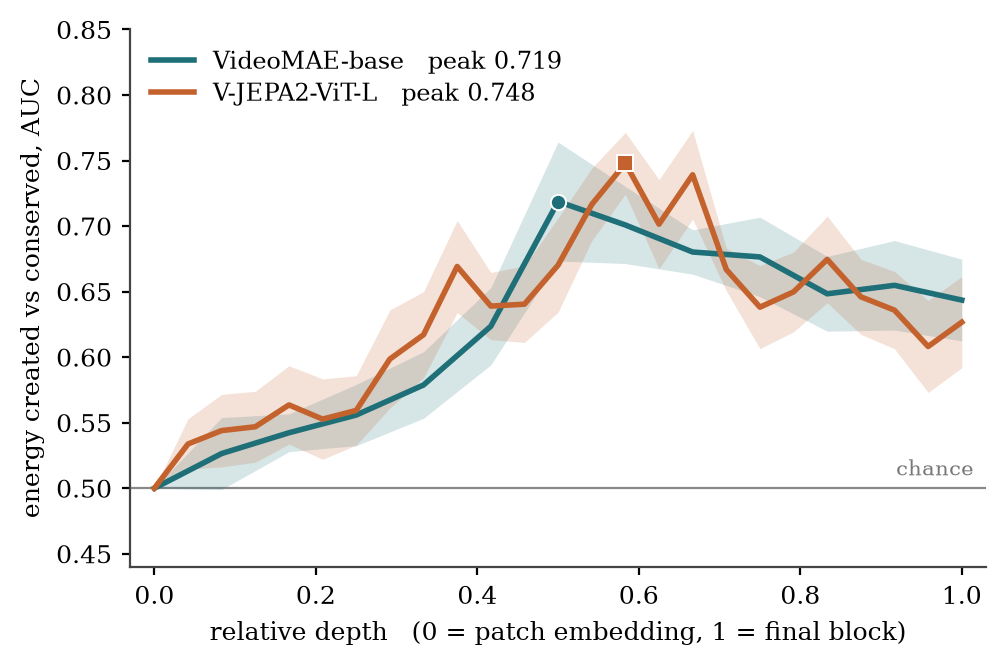}
\end{minipage}
\hfill
\begin{minipage}[b]{0.56\linewidth}
  \centering
  \includegraphics[height=3.3cm]{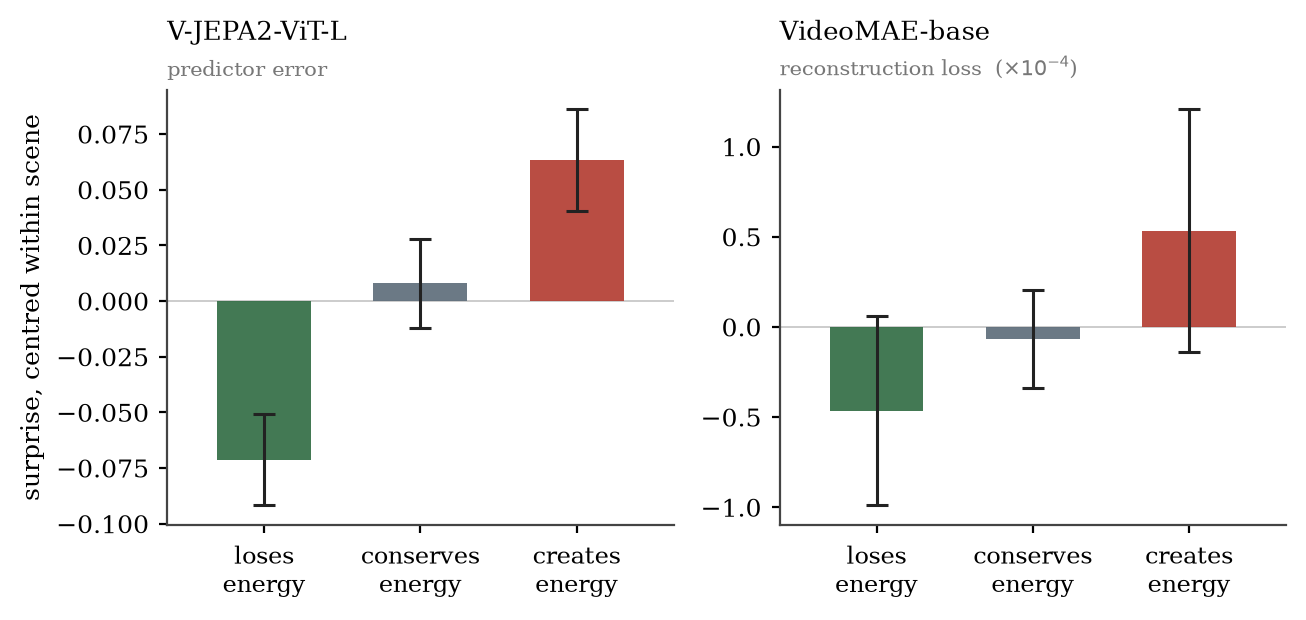}
\end{minipage}
\caption{\textbf{Energy created against energy conserved, read two ways.}
\emph{Left:} out-of-fold AUC of a linear probe separating the clips that create energy
($e = k$) from those that conserve it ($e = 1$) across depth, pooled \texttt{all.pre}; momentum is conserved in both sets. Both encoders
peak far below an observer given the contact geometry, which reads the same label at
$1.000$. \emph{Right:} the models' own surprise, with nothing fitted. Each is shown only the
pre-contact segment, which the three endings share exactly, and emits one forecast. Note the differing $y$ scales. }
\label{fig:energy}
\end{figure}

\paragraph{Probing energy creation.}

We first test whether the encoders distinguish collisions that create kinetic energy from those that lose it. The resulting probes achieve AUCs of ($0.719 \pm 0.045$) for VideoMAE and ($0.748 \pm 0.024$) for V-JEPA2. Both outperform kinematic observers that lack contact geometry, whose AUCs range from $0.52$--$0.63$, but fall short of an observer given the contact geometry, which achieves an AUC of $\mathbf{1.000}$. This comparison shows that energy creation is decodable from the representations, but does not establish that the encoders explicitly represent kinetic energy: the binary label is also recoverable from visible kinematics through the coefficient of restitution. We therefore turn to two complementary tests: whether the models' native pretraining objectives respond systematically to different collision outcomes, and whether a kinetic-energy readout transfers across scenarios.

\paragraph{Reading the predictor.}
We next examine the models' native pretraining objectives using three collision outcomes that share identical pre-contact frames: an elastic collision that conserves kinetic energy ($e=1$), an inelastic collision that loses kinetic energy ($e=0.15$), and an energy-creating collision that violates energy conservation ($e=1.5$).
V-JEPA2 uses a predictor to forecast representations of future frames.
Given only the pre-contact segment, it produces a single forecast, which we compare with the encoded representation of each outcome.
VideoMAE uses a pretraining decoder to reconstruct pixels: we mask the post-contact frames and measure reconstruction loss, using identical mask draws across the three outcomes.
We restrict the comparison to the regions occupied by the spheres, since the three endings are pixel-identical elsewhere.
V-JEPA2 finds the energy-creating outcome more surprising than the energy-losing outcome in 95\% of scenes under both L1 and squared error ($p < 0.0001$), and in 90\% under cosine distance ($p = 0.0002$).
VideoMAE shows the same trend, but the difference is not statistically significant: the energy-creating outcome produces greater reconstruction loss in 65\% of scenes ($p = 0.132$).
These results suggest that V-JEPA2's predictions are systematically sensitive to the collision outcome, while the evidence from VideoMAE is inconclusive.

\paragraph{Does the energy readout survive a change of scenario?}
Everything above is measured within one family of scenes. We test transfer directly against
a second set in which a single cube decelerates to rest on a friction floor (see Section~\ref{sec:friction} for detailed description), so kinetic
energy is lost to friction rather than to deformation. A readout is fitted on the
sliding cube dataset, then applied unchanged to the
collisions; we score by Spearman $\rho$, since the two domains put the quantity on different
scales. Both encoders read kinetic energy well within the source set ($0.971$ and $0.950$).
Transferred, VideoMAE lands at chance ($\rho = 0.022$, $p = 0.725$) while V-JEPA2 reaches
$\rho = 0.275$ ($p < 0.001$). We also train two probes for speed and size of the object on the slide dataset and transfer to the collision dataset: size transfers at $0.631$ and $0.585$, and speed
at $0.244$ and $0.428$. Transferring size and speed \emph{separately} and forming
$\mathrm{size}\times\mathrm{speed}^2$ from the two transferred predictions reaches $0.291$
and $0.324$, above the  direct kinetic energy readout on both models. Thus, what survives the change in environment are the physical factors;
kinetic energy is assembled at readout and does not generalize.

\section{Friction}
\label{sec:friction}

\begin{table}[t]
\centering
\small
\caption{\textbf{Reading $\mu$ directly and composing it from decoded $(a,\theta)$ give
the same answer, because on this set they are the same function.} Out-of-fold $R^2$ on the
incline set ($n=150$), grouped folds with nested ridge penalty, mean $\pm$ sd over $12$
fold seeds; each encoder at its own best pool. \textsc{direct} regresses $\mu$ from
activations. \textsc{composed} decodes $a$ and $\theta$ separately and learns a
second-stage linear map from the out-of-fold $(\hat a, \hat\theta)$ to $\mu$. Every stage
is out-of-fold, so no scene's own $\hat a$ or $\hat\theta$ is used to fit the map that
scores it. Over the sampled range $\mu$ is $0.997$ linear in $(a,\theta)$, so the two
routes are the same function and the tie is uninformative rather than positive.}
\label{tab:composed-mu}
\begin{tabular}{lcc}
\toprule
& \multicolumn{2}{c}{$\mu$, out-of-fold $R^2$} \\
\cmidrule(lr){2-3}
Model & \textsc{direct} & \textsc{composed} \\
\midrule
VideoMAE-base & $0.512 \pm 0.037$ & $0.498 \pm 0.034$ \\
V-JEPA2-ViT-L & $0.765 \pm 0.013$ & $0.759 \pm 0.018$ \\
\bottomrule
\end{tabular}
\end{table}

\begin{table}[t]
\centering
\small
\caption{\textbf{A friction coefficient decoded within one floor geometry does not survive
a change of geometry, while cube size does.} Probes are trained on one dataset and applied
unchanged to the other, restricted to the overlapping $\mu$ range $[0.151, 0.399]$ ($164$
slide / $148$ incline scenes), pool \texttt{all.all}, each domain standardised by its own
moments so only the \emph{direction} has to transfer. The two sets make $\mu$ a different
function of the visible kinematics --- flat: $\mu = a/g$ with
$\mathrm{corr}(\mu,a) = +0.9998$; tilted: $\mu = (\sin\theta - a/g)/\cos\theta$ with
$\mathrm{corr}(\mu,a) = -0.645$ --- so a readout that is really tracking acceleration
cannot cross. \textsc{size} is the control: it faces the identical domain gap (camera $14$
vs $18$, flat vs tilted, different band pattern). The third row of each block is the
diagnostic: the same $\mu$ probe scored against the target set's \emph{acceleration}.}
\label{tab:transfer}
\begin{tabular}{llcrr}
\toprule
Model & Readout & Within-source $R^2$ & Transfer $R^2$ & Transfer corr. \\
\midrule
\multicolumn{5}{l}{\emph{Train on incline $\rightarrow$ test on slide}} \\
\addlinespace[1pt]
VideoMAE-base & $\mu$                        & 0.127 & $-0.663$ & $+0.541$ \\
VideoMAE-base & size \emph{(control)}        & 0.982 & $\mathbf{+0.829}$ & $\mathbf{+0.939}$ \\
VideoMAE-base & $\mu$ probe $\rightarrow a$  & 0.127 & $-0.666$ & $+0.540$ \\
\addlinespace[2pt]
V-JEPA2-ViT-L & $\mu$                        & 0.622 & $-0.925$ & $-0.072$ \\
V-JEPA2-ViT-L & size \emph{(control)}        & 0.982 & $\mathbf{+0.874}$ & $\mathbf{+0.948}$ \\
V-JEPA2-ViT-L & $\mu$ probe $\rightarrow a$  & 0.622 & $-0.926$ & $-0.073$ \\
\midrule
\multicolumn{5}{l}{\emph{Train on slide $\rightarrow$ test on incline}} \\
\addlinespace[1pt]
VideoMAE-base & $\mu$                        & 0.937 & $-0.167$ & $+0.072$ \\
VideoMAE-base & size \emph{(control)}        & 0.995 & $+0.172$ & $+0.440$ \\
VideoMAE-base & $\mu$ probe $\rightarrow a$  & 0.937 & $\mathbf{+0.262}$ & $\mathbf{+0.513}$ \\
\addlinespace[2pt]
V-JEPA2-ViT-L & $\mu$                        & 0.944 & $-0.553$ & $-0.176$ \\
V-JEPA2-ViT-L & size \emph{(control)}        & 0.990 & $+0.449$ & $+0.826$ \\
V-JEPA2-ViT-L & $\mu$ probe $\rightarrow a$  & 0.944 & $\mathbf{+0.586}$ & $\mathbf{+0.793}$ \\
\bottomrule
\end{tabular}
\end{table}

\begin{figure}[t]
\centering
\includegraphics[width=0.70\linewidth]{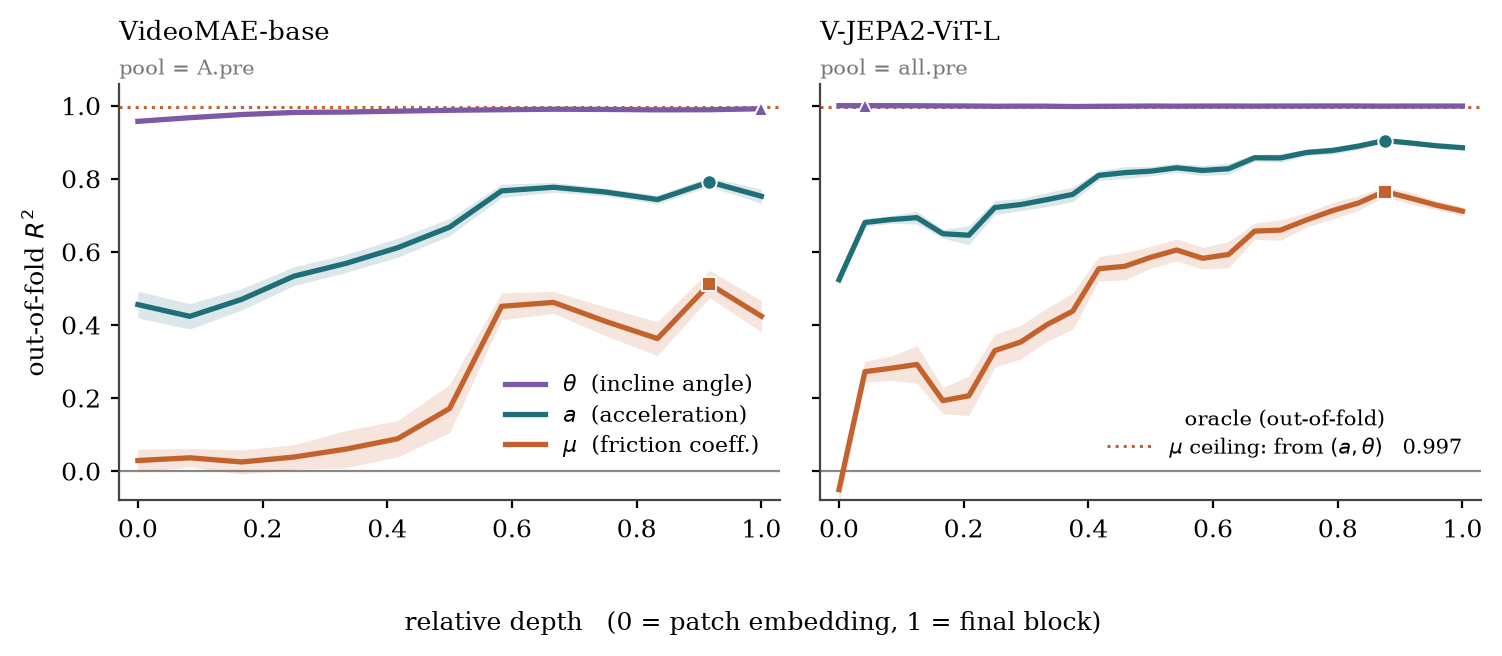}
\caption{\textbf{The ingredients of friction decode; the coefficient does not follow.}
Out-of-fold $R^2$ for incline angle $\theta$, down-slope acceleration $a$ and friction
coefficient $\mu$ across depth on $150$ incline scenes, mean over $12$ fold seeds with
$\pm1$ sd bands. Each encoder is shown at its own best pool. The dotted line is the oracle
ceiling: $\mu$ recovered from $(a,\theta)$ on ground-truth labels through the identical
protocol, $R^2 = 0.997$. $\theta$ is already at $0.957$ and $1.000$ at the patch
embedding.}
\label{fig:friction}
\end{figure}

\paragraph{Dataset.}
A cube is released on an inclined plane, with the incline angle
$\theta$ and friction coefficient $\mu$ sampled independently
for each of 150 scenes. The down-slope acceleration $a$ is
measured from the cube's trajectory. We use the known values of
$\theta$ and $\mu$, together with the measured acceleration $a$,
to evaluate how well the encoders' representations support
decoding these quantities. We also construct a companion dataset
in which the same cube slides on a flat, banded floor, where
$\mu = a/g$. Unlike the collision experiments, neither setting
contains a discrete contact event that separates the video into
pre- and post-contact phases. The temporal windows defined in
\S\ref{sec:prelim} therefore correspond to early and late portions of
the sliding video rather than pre- and post-contact frames.

\paragraph{Decoding friction.}
We train separate probes on the encoder representations to predict
the incline angle $\theta$, down-slope acceleration $a$, and
friction coefficient $\mu$ for the inclined-plane scenes (Fig.~4).
Both encoders recover the incline angle and acceleration more
accurately than the friction coefficient. The best $R^2$ values
for $\theta$ are 0.991 for VideoMAE-base and 1.000 for
V-JEPA2-ViT-L. These values are already 0.957 and 1.000,
respectively, at the patch-embedding layer, suggesting that the
incline angle is largely accessible from static scene geometry.
The best acceleration readouts reach $R^2=0.791\pm0.013$ and
$0.905\pm0.006$, respectively, while direct readouts of $\mu$
reach only $0.512\pm0.037$ and $0.765\pm0.013$. To assess how
well friction can be recovered from these quantities, we fit
observers using ground-truth values. An observer given $a$ alone
predicts $\mu$ with $R^2=0.402$, whereas one given both $a$ and
$\theta$ reaches $R^2=0.997$. Thus, although acceleration and
incline angle are highly decodable, direct friction readouts
do not match the accuracy achievable from their combination.

\paragraph{Composing friction from decoded quantities.}
We compare the direct friction readout with an estimate obtained
by combining separately decoded acceleration $\hat{a}$ and incline
angle $\hat{\theta}$ (Table~1). The two approaches achieve similar
performance: for VideoMAE-base, the direct and composed readouts
reach $R^2=0.512\pm0.037$ and $0.498\pm0.034$, respectively;
for V-JEPA2-ViT-L, they reach $R^2=0.765\pm0.013$ and
$0.759\pm0.018$. However, this similarity does not establish
whether friction is directly accessible from the encoder
representations or can be recovered through decoded acceleration
and incline angle. Over the sampled range, a linear model using
ground-truth acceleration and incline angle predicts friction
with $R^2=0.997$, showing that these quantities together provide
nearly all the information needed for this prediction. Because
the direct and composed readouts perform similarly, this
comparison cannot distinguish between the two explanations.

\paragraph{Transfer across floor geometries.}
We next test whether a friction readout generalizes between the
inclined-plane and flat-floor datasets (Table~2). The relationship
between friction and acceleration differs across these geometries:
on the flat floor, $\mu=a/g$, whereas on the incline,
$\mu=(g\sin\theta-a)/(g\cos\theta)$, where $a$ is the signed
acceleration along the slope. Consequently, a readout that relies
on acceleration in one geometry need not predict friction
correctly in the other. We train each probe on one geometry and
apply it unchanged to the other, restricting evaluation to the
overlapping friction-coefficient range and standardizing each
domain using its own moments. Friction readouts fail to transfer
in both directions for both encoders. When trained on the incline
and tested on the flat floor, transfer $R^2$ is $-0.663$ for
VideoMAE-base and $-0.925$ for V-JEPA2-ViT-L. In the reverse
direction, transfer $R^2$ is $-0.167$ and $-0.553$, respectively.
By contrast, cube size transfers substantially better, with
correlations of $0.939$ and $0.948$ for incline-to-flat transfer,
and $0.440$ and $0.826$ for flat-to-incline transfer. This
control shows that the geometry shift does not prevent all
properties from generalizing.

As a further diagnostic, we evaluate whether a friction probe
trained on one geometry predicts acceleration rather than friction
when applied to the other. For probes trained on the flat floor
and tested on the incline, the predictions explain more variance
in acceleration ($R^2=0.262$ for VideoMAE-base and $0.586$ for
V-JEPA2-ViT-L) than in friction itself ($R^2=-0.167$ and
$-0.553$, respectively). This pattern is consistent with the
probes relying on geometry-dependent kinematic cues rather than
a friction readout that generalizes across geometries. It does
not establish that the encoders contain no information about
friction; rather, the tested friction readouts do not transfer
reliably between these two settings.
\vspace{-5pt}

\section{Conclusion}
\label{sec:conclusion}

We set out to ask whether video encoders hold physical quantities that are defined between
objects, and our early findings are mixed. Both encoders separate elastic from inelastic
collisions with high reliability (AUC $0.911$ and $0.973$), but the label there is a
mass-free ratio of two visible speeds, so this establishes that they represent how a
collision \emph{looked}, not its kinetic energy. Asked without any fitted readout, V-JEPA2's
own predictor is reliably more surprised by a collision that \emph{creates} energy than by
one that loses it, and the elastic control rules out the account that it is merely tracking
departure from elastic. Momentum-violation detection, by contrast, appears in only one of
the two encoders, and neither model holds a coefficient of friction that survives a change
of floor geometry.

There are a number of limitations in this early exploration. The stimuli are synthetic and
simple. Two encoders cannot tell us whether the momentum split reflects the training
objective or the scale. The energy finding rests on a single restitution ratio and on one
model's predictor. And decodability is not use: a quantity absent from a linear readout may
still be computed nonlinearly downstream. We intend to add
models from both families and move toward stimuli in which the aftermath is only partly
observed.

\bibliographystyle{plainnat}
\bibliography{bib}

\begin{thebibliography}{15}
\providecommand{\natexlab}[1]{#1}
\providecommand{\url}[1]{\texttt{#1}}
\expandafter\ifx\csname urlstyle\endcsname\relax
  \providecommand{\doi}[1]{doi: #1}\else
  \providecommand{\doi}{doi: \begingroup \urlstyle{rm}\Url}\fi

\bibitem[Alam(2026)]{alam2026steering}
Nahid Alam.
\newblock Causal physics steering in video world models via concept activation vectors, 2026.
\newblock URL \url{https://arxiv.org/abs/2605.24322}.

\bibitem[Assran et~al.(2025)Assran, Bardes, Fan, Garrido, Howes, Komeili, Muckley, Rizvi, Roberts, Sinha, Zholus, Arnaud, Gejji, Martin, Hogan, Dugas, Bojanowski, Khalidov, Labatut, Massa, Szafraniec, Krishnakumar, Li, Ma, Chandar, Meier, LeCun, Rabbat, and Ballas]{assran2025vjepa2}
Mido Assran, Adrien Bardes, David Fan, Quentin Garrido, Russell Howes, Mojtaba Komeili, Matthew Muckley, Ammar Rizvi, Claire Roberts, Koustuv Sinha, Artem Zholus, Sergio Arnaud, Abha Gejji, Ada Martin, Francois~Robert Hogan, Daniel Dugas, Piotr Bojanowski, Vasil Khalidov, Patrick Labatut, Francisco Massa, Marc Szafraniec, Kapil Krishnakumar, Yong Li, Xiaodong Ma, Sarath Chandar, Franziska Meier, Yann LeCun, Michael Rabbat, and Nicolas Ballas.
\newblock V-jepa 2: Self-supervised video models enable understanding, prediction and planning, 2025.
\newblock URL \url{https://arxiv.org/abs/2506.09985}.

\bibitem[Bear et~al.(2021)Bear, Wang, Mrowca, Binder, Tung, Pramod, Holdaway, Tao, Smith, Sun, Li, Kanwisher, Tenenbaum, Yamins, and Fan]{bear2021physion}
Daniel Bear, Elias Wang, Damian Mrowca, Felix~J. Binder, Hsiao{-}Yu Tung, R.~T. Pramod, Cameron Holdaway, Sirui Tao, Kevin~A. Smith, Fan{-}Yun Sun, Fei{-}Fei Li, Nancy Kanwisher, Josh Tenenbaum, Dan Yamins, and Judith~E. Fan.
\newblock Physion: Evaluating physical prediction from vision in humans and machines.
\newblock In Joaquin Vanschoren and Sai{-}Kit Yeung, editors, \emph{Proceedings of the Neural Information Processing Systems Track on Datasets and Benchmarks 1, NeurIPS Datasets and Benchmarks 2021, December 2021, virtual}, 2021.
\newblock URL \url{https://datasets-benchmarks-proceedings.neurips.cc/paper/2021/hash/d09bf41544a3365a46c9077ebb5e35c3-Abstract-round1.html}.

\bibitem[Belinkov(2022)]{belinkov2022probing}
Yonatan Belinkov.
\newblock Probing classifiers: Promises, shortcomings, and advances.
\newblock \emph{Computational Linguistics}, 48\penalty0 (1):\penalty0 207--219, March 2022.
\newblock \doi{10.1162/coli\_a\_00422}.

\bibitem[Esmati et~al.(2026)Esmati, Nath, Hofmann, Nowrouzezahrai, Kahou, and Mirmehdi]{esmati2026invisible}
Parsa Esmati, Somjit Nath, Katja Hofmann, Derek Nowrouzezahrai, Samira~Ebrahimi Kahou, and Majid Mirmehdi.
\newblock The invisible hand of physics: When video diffusion models know more than they show, 2026.
\newblock URL \url{https://arxiv.org/abs/2606.05328}.

\bibitem[Garrido et~al.(2025)Garrido, Ballas, Assran, Bardes, Najman, Rabbat, Dupoux, and LeCun]{garrido2025intuitive}
Quentin Garrido, Nicolas Ballas, Mahmoud Assran, Adrien Bardes, Laurent Najman, Michael Rabbat, Emmanuel Dupoux, and Yann LeCun.
\newblock Intuitive physics understanding emerges from self-supervised pretraining on natural videos, 2025.
\newblock URL \url{https://arxiv.org/abs/2502.11831}.

\bibitem[Greff et~al.(2022)Greff, Belletti, Beyer, Doersch, Du, Duckworth, Fleet, Gnanapragasam, Golemo, Herrmann, Kipf, Kundu, Lagun, Laradji, Liu, Meyer, Miao, Nowrouzezahrai, {\"{O}}ztireli, Pot, Radwan, Rebain, Sabour, Sajjadi, Sela, Sitzmann, Stone, Sun, Vora, Wang, Wu, Yi, Zhong, and Tagliasacchi]{greff2022kubric}
Klaus Greff, Francois Belletti, Lucas Beyer, Carl Doersch, Yilun Du, Daniel Duckworth, David~J. Fleet, Dan Gnanapragasam, Florian Golemo, Charles Herrmann, Thomas Kipf, Abhijit Kundu, Dmitry Lagun, Issam~H. Laradji, Hsueh{-}Ti~Derek Liu, Henning Meyer, Yishu Miao, Derek Nowrouzezahrai, A.~Cengiz {\"{O}}ztireli, Etienne Pot, Noha Radwan, Daniel Rebain, Sara Sabour, Mehdi S.~M. Sajjadi, Matan Sela, Vincent Sitzmann, Austin Stone, Deqing Sun, Suhani Vora, Ziyu Wang, Tianhao Wu, Kwang~Moo Yi, Fangcheng Zhong, and Andrea Tagliasacchi.
\newblock Kubric: {A} scalable dataset generator.
\newblock In \emph{{IEEE/CVF} Conference on Computer Vision and Pattern Recognition, {CVPR} 2022, New Orleans, LA, USA, June 18-24, 2022}, pages 3739--3751. {IEEE}, 2022.
\newblock \doi{10.1109/CVPR52688.2022.00373}.

\bibitem[Hewitt and Liang(2019)]{hewitt2019control}
John Hewitt and Percy Liang.
\newblock Designing and interpreting probes with control tasks.
\newblock In Kentaro Inui, Jing Jiang, Vincent Ng, and Xiaojun Wan, editors, \emph{Proceedings of the 2019 Conference on Empirical Methods in Natural Language Processing and the 9th International Joint Conference on Natural Language Processing (EMNLP-IJCNLP)}, pages 2733--2743, Hong Kong, China, November 2019. Association for Computational Linguistics.
\newblock \doi{10.18653/v1/D19-1275}.

\bibitem[Joseph et~al.(2026)Joseph, Garrido, Balestriero, Kowal, Fel, Bakhtiari, Richards, and Rabbat]{joseph2026interpreting}
Sonia Joseph, Quentin Garrido, Randall Balestriero, Matthew Kowal, Thomas Fel, Shahab Bakhtiari, Blake Richards, and Mike Rabbat.
\newblock Interpreting physics in video world models, 2026.
\newblock URL \url{https://arxiv.org/abs/2602.07050}.

\bibitem[Kim et~al.(2018)Kim, Wattenberg, Gilmer, Cai, Wexler, Vi{\'{e}}gas, and Sayres]{kim2018tcav}
Been Kim, Martin Wattenberg, Justin Gilmer, Carrie~J. Cai, James Wexler, Fernanda~B. Vi{\'{e}}gas, and Rory Sayres.
\newblock Interpretability beyond feature attribution: Quantitative testing with concept activation vectors {(TCAV)}.
\newblock In Jennifer~G. Dy and Andreas Krause, editors, \emph{Proceedings of the 35th International Conference on Machine Learning, {ICML} 2018, Stockholmsm{\"{a}}ssan, Stockholm, Sweden, July 10-15, 2018}, volume~80 of \emph{Proceedings of Machine Learning Research}, pages 2673--2682. {PMLR}, 2018.
\newblock URL \url{http://proceedings.mlr.press/v80/kim18d.html}.

\bibitem[Punzo et~al.(2026)Punzo, Caselli, Pantelidis, Massafra, Sardo, and Salehi]{punzo2026layerwise}
Samuele Punzo, Niccol{\`o} Caselli, Ippokratis Pantelidis, Francesco Massafra, Salvatore~Lo Sardo, and Mohammadreza Salehi.
\newblock Do video foundation models understand intuitive physics? a layerwise probing analysis, 2026.
\newblock URL \url{https://arxiv.org/abs/2606.09646}.

\bibitem[Riochet et~al.(2020)Riochet, Castro, Bernard, Lerer, Fergus, Izard, and Dupoux]{riochet2018intphys}
Ronan Riochet, Mario~Ynocente Castro, Mathieu Bernard, Adam Lerer, Rob Fergus, V{\'e}ronique Izard, and Emmanuel Dupoux.
\newblock Intphys: A framework and benchmark for visual intuitive physics reasoning, 2020.
\newblock URL \url{https://arxiv.org/abs/1803.07616}.

\bibitem[Tong et~al.(2022)Tong, Song, Wang, and Wang]{tong2022videomae}
Zhan Tong, Yibing Song, Jue Wang, and Limin Wang.
\newblock Videomae: Masked autoencoders are data-efficient learners for self-supervised video pre-training.
\newblock In Sanmi Koyejo, S.~Mohamed, A.~Agarwal, Danielle Belgrave, K.~Cho, and A.~Oh, editors, \emph{Advances in Neural Information Processing Systems 35: Annual Conference on Neural Information Processing Systems 2022, NeurIPS 2022, New Orleans, LA, USA, November 28 - December 9, 2022}, 2022.
\newblock URL \url{http://papers.nips.cc/paper\_files/paper/2022/hash/416f9cb3276121c42eebb86352a4354a-Abstract-Conference.html}.

\bibitem[Upadhyay et~al.(2026)Upadhyay, Zhang, Solomon, Agrawal, Ba, Wong, de~Melo, and Kadambi]{upadhyay2026worldbench}
Rishi Upadhyay, Howard Zhang, Jim Solomon, Ayush Agrawal, Yunhao Ba, Alex Wong, Celso~M de~Melo, and Achuta Kadambi.
\newblock Worldbench: Benchmarking physical understanding of world models by isolating physics concepts, 2026.
\newblock URL \url{https://arxiv.org/abs/2601.21282}.

\bibitem[Yi et~al.(2020)Yi, Gan, Li, Kohli, Wu, Torralba, and Tenenbaum]{yi2020clevrer}
Kexin Yi, Chuang Gan, Yunzhu Li, Pushmeet Kohli, Jiajun Wu, Antonio Torralba, and Joshua~B. Tenenbaum.
\newblock {CLEVRER:} collision events for video representation and reasoning.
\newblock In \emph{8th International Conference on Learning Representations, {ICLR} 2020, Addis Ababa, Ethiopia, April 26-30, 2020}. OpenReview.net, 2020.
\newblock URL \url{https://openreview.net/forum?id=HkxYzANYDB}.

\end{thebibliography}


\appendix

\section{Physical Attributes in Multi-Object Interaction Videos}\label{sec:single}

\begin{figure}[ht]
\centering
\includegraphics[width=\linewidth]{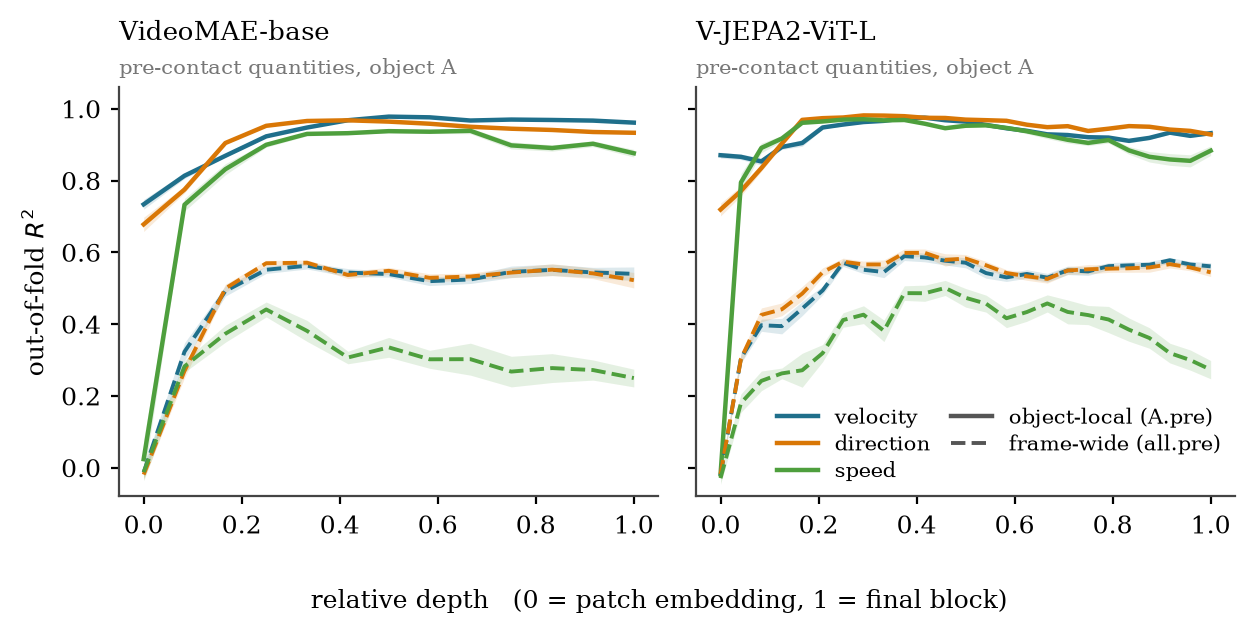}
\caption{\textbf{Whether attribute emergence is visible depends on the pooling, not on the
model.} Out-of-fold $R^2$ across depth on the $150$ momentum-conserving clips, mean over
$12$ fold seeds with $\pm1$ sd bands. \emph{Solid:} object-local pooling, object A's own
tokens. \emph{Dashed:} frame-wide pooling, the protocol of the single-object literature.
Under object-local pooling the quantities are already available at the patch embedding and
nothing emerges; under frame-wide pooling every curve starts at zero and rises through the
first third. Mask-only floors at \texttt{A.pre}, computed under the identical protocol, are
negligible ($-0.37$ to $+0.05$ for VideoMAE-base, $-0.15$ to $+0.22$ for V-JEPA2-ViT-L).}
\label{fig:single-object}
\end{figure}

\begin{figure}[ht]
\centering
\includegraphics[width=\linewidth]{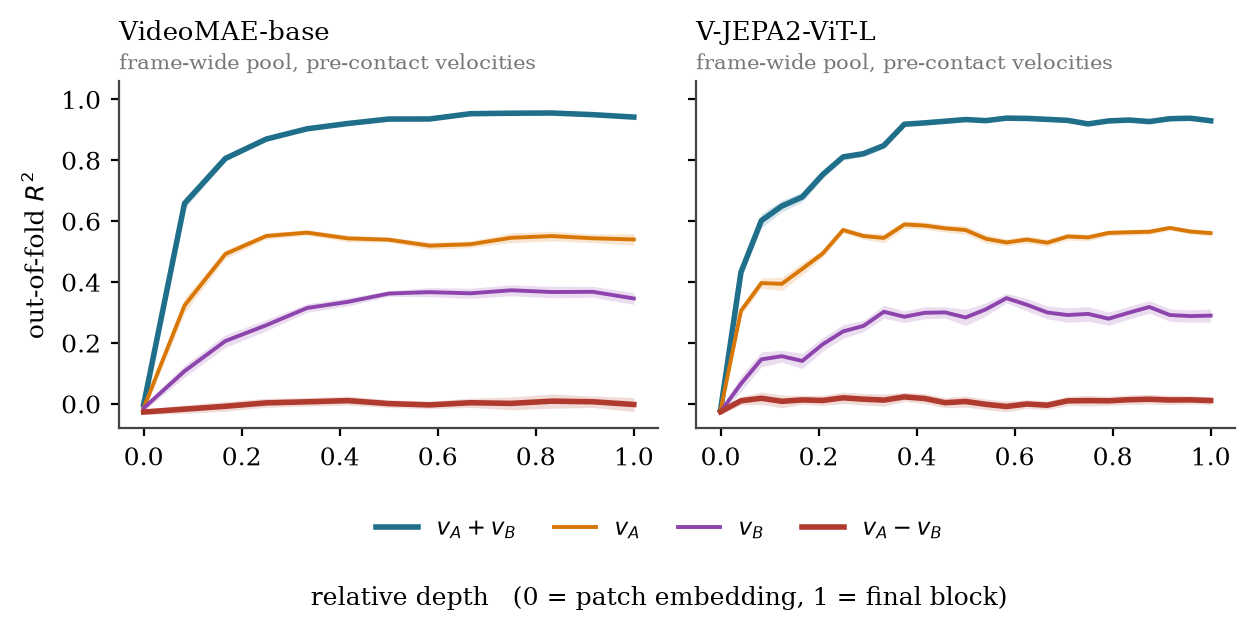}
\caption{\textbf{A frame-wide pool carries the sum of the two objects and not their
difference.} Out-of-fold $R^2$ across depth, pool \texttt{all.pre}, $12$ fold seeds with
$\pm1$ sd bands. The symmetric combination $v_A + v_B$ reaches $0.955$ and $0.938$, well
above either individual velocity, while the antisymmetric combination $v_A - v_B$ peaks at
$0.011$ and $0.023$---those are maxima over all $13$ and $25$ layers, so there is no depth
at which the difference is recoverable.}
\label{fig:pool-sum}
\end{figure}

\begin{table}[ht]
\centering
\small
\caption{\textbf{The same four quantities, read from four pools.} Out-of-fold $R^2$ at the
best layer. Object-local pools read their own object well and the other poorly; the
frame-wide pools reverse this, reading each object worse than its own pool does while
reading the sum better than any object-local pool reads anything.}
\label{tab:pool-decomp}
\begin{tabular}{llcccc}
\toprule
Model & Pool & $v_A$ & $v_B$ & $v_A + v_B$ & $v_A - v_B$ \\
\midrule
VideoMAE-base & \texttt{A.pre}   & $\mathbf{0.978}$ & $0.739$ & $0.865$ & $0.887$ \\
              & \texttt{B.pre}   & $0.794$ & $\mathbf{0.960}$ & $0.861$ & $0.861$ \\
              & \texttt{all.pre} & $0.563$ & $0.373$ & $\mathbf{0.955}$ & $0.011$ \\
              & \texttt{all.all} & $0.519$ & $0.350$ & $0.920$ & $0.005$ \\
\midrule
V-JEPA2-ViT-L & \texttt{A.pre}   & $\mathbf{0.975}$ & $0.729$ & $0.845$ & $0.876$ \\
              & \texttt{B.pre}   & $0.779$ & $\mathbf{0.953}$ & $0.834$ & $0.868$ \\
              & \texttt{all.pre} & $0.589$ & $0.347$ & $\mathbf{0.938}$ & $0.023$ \\
              & \texttt{all.all} & $0.558$ & $0.322$ & $0.901$ & $0.018$ \\
\bottomrule
\end{tabular}
\end{table}

\paragraph{The ingredients are present.}
Read from its own token pool, every per-object attribute we probe reaches $0.92$--$0.98$:
velocity at $0.978$ and $0.975$, direction at $0.968$ and $0.981$, speed at $0.938$ and
$0.971$, radius at $0.922$ and $0.966$, and log mass at $0.952$ and $0.982$, for
VideoMAE-base and V-JEPA2-ViT-L. Mask-only floors under the identical protocol are
negligible, so object-local pooling is not carrying these numbers. This is the baseline the
main text needs: the relational quantities that fail there fail with their constituents
present near ceiling, not absent.

\paragraph{Pooling decides whether emergence is visible.}
The same probes under the two poolings give qualitatively different pictures
(Fig.~\ref{fig:single-object}). Object-locally, velocity and direction are already at
$0.73/0.87$ and $0.68/0.72$ at the patch embedding, before any block has run, so the curves
are flat and there is no transition to observe. Frame-wide, every quantity starts at zero
and rises to its peak within roughly the first third of the network---the depth profile
reported for single-object stimuli, including the decay toward the output layers that is
clearest in VideoMAE's speed curve ($0.441$ at one quarter depth, $0.25$ at the final
block). We do not, however, reproduce the asymmetry that \citet{joseph2026interpreting}
identify as characterising that transition: they report scalar magnitudes available early
and direction becoming decodable only at the transition, whereas in our data direction
rises at least as fast as speed in both models.

We think the two-object design is why, and that neither pooling is a clean replication of a
single-object protocol. Object-local pooling is \emph{cleaner} than a frame-wide mean, since
mask selection discards background tokens, which is why the quantities are linearly
available before any block runs. Frame-wide pooling is \emph{dirtier}, because with two
spheres in frame the pool holds a superposition; this is why the dashed curves cap near
$0.57$ rather than the $0.95$ the same quantities reach object-locally. For a single moving
object the two coincide and neither effect arises.

\paragraph{A frame-wide pool computes a sum.}
That superposition is measurable directly, and it is what connects this appendix to
\S\ref{sec:momentum}. Reading all four combinations across depth
(Fig.~\ref{fig:pool-sum}), the symmetric combination $v_A + v_B$ reaches $0.955$ and
$0.938$---above either individual velocity, and above what any object-local pool achieves
for it (Table~\ref{tab:pool-decomp})---while the antisymmetric combination $v_A - v_B$
peaks at $0.011$ and $0.023$ across every layer of both models. A pool that merely degraded
both objects would place the sum between the parts rather than far above them. What a
frame-wide mean does is superpose an object-agnostic per-token code, and a sum cannot be
un-mixed.

Momentum transfer $\Delta p = m_A \Delta v_A + m_B \Delta v_B$ is a sum over the two
objects, so it belongs to the combination this pooling preserves. That is why it decodes
frame-wide ($R^2 = +0.426$) and not from the object-local pools that read its own
ingredients far better ($-0.404$ concatenated). The pooling geometry, not the physics,
decides which relational quantities a linear readout can reach.

\section{Extended Related Work}
\label{sec:related}

\paragraph{Behavioral evaluation of physical understanding.}
Most evidence about physics in video models comes from benchmarks that score an outcome.
CLEVRER \citep{yi2020clevrer} poses descriptive, predictive, explanatory and
counterfactual questions over short videos of colliding objects; Physion
\citep{bear2021physion} asks whether two designated objects will make contact across eight
scenario families and compares model predictions with human judgements; IntPhys
\citep{riochet2018intphys} uses a violation-of-expectation design in which a model must
prefer the possible continuation over the impossible one; and WorldBench
\citep{upadhyay2026worldbench} isolates individual physics concepts for diagnostic
scoring. \citet{garrido2025intuitive} apply the violation-of-expectation framing to
representation learning and find that intuitive-physics competence emerges from
self-supervised video pretraining, while pixel-space prediction performs near chance, and
V-JEPA~2 \citep{assran2025vjepa2} reports $98\%$ zero-shot accuracy on IntPhys. These
studies establish that competent behavior exists; by construction they do not identify
what internal quantity produces it, and a representation tracking a visible correlate is
scored identically to one encoding the underlying variable.

\paragraph{Interpretability of video encoders.}
A newer line examines video encoders through mechanistic interpretability. \citet{joseph2026interpreting} give the first
interpretability study of physical variables inside large-scale video encoders. Probing
V-JEPA~2 and VideoMAE-v2 layer by layer, they locate a sharp mid-depth transition---the
``Physics Emergence Zone''---at which possible-versus-impossible discrimination and motion
direction both become decodable, show that scalar magnitudes such as speed are available
much earlier, and characterise direction as a high-dimensional circular population code
requiring coordinated multi-feature intervention to steer. They further show that the
direction and plausibility subspaces are nearly orthogonal, ruling out representational
reuse, and conclude against physics-engine-style compact latent state in favour of
distributed, task-specific representations. \citet{punzo2026layerwise} run layerwise
frozen-feature probes across V-JEPA, VideoMAE and a diffusion generator.
\citet{esmati2026invisible} extend the approach to video diffusion models by inverting the
sampler to recover latent trajectories for real videos, and find plausibility linearly
decodable inside the denoiser though absent from the VAE latent it operates on. Finally,
\citet{alam2026steering} treats a probe's weight vector as a concept activation vector
\citep{kim2018tcav} and injects it at inference, shifting a plausibility judgement in
either direction and localising the effect to the same mid-depth band.

\paragraph{What remains unexamined.}
These studies probe \emph{per-object} quantities---position, speed, direction,
acceleration---or a single binary plausibility bit. Multi-object stimuli do appear:
\citet{joseph2026interpreting} verify on CLEVRER that the emergence signature for
direction survives with object-level probes, and use CLEVRER object counting as a
\emph{negative} control showing the signature is not a generic property of video
processing. But in both cases the target is a property of one object, or of the clip as a
whole, and never a quantity defined \emph{between} objects. Momentum transfer $\Delta p$,
restitution $e$, and a friction coefficient $\mu$ cannot be read off either body in
isolation; they are relations, and they are the quantities a world model would need in
order to generalise across scenes rather than memorise them. To our knowledge no prior
work measures them at the representational level, which is what we do here.

\paragraph{Probing methodology.}
Our evaluation protocol follows a line of work in NLP interpretability arguing that a
probe's accuracy is uninterpretable without a matched baseline. \citet{hewitt2019control}
introduce control tasks---structureless targets fitted through the identical probe
family---to separate what a representation encodes from what a probe can learn, and
\citet{belinkov2022probing} surveys the resulting methodological corrections. We adapt the
same discipline to physics: rather than a randomised control target, each probe is scored
against an observer given the visible scalars a camera supplies and fitted through the
identical pipeline and folds, since the physical relations of interest are frequently
algebraic functions of those scalars.


\section{Definitions of the Probed Quantities}
\label{sec:variables}

Every target is taken from the simulator's recorded state rather than from the pixels, so
no target carries measurement error. Collisions are frictionless and gravity is zero on the
collision set, so both spheres move in a plane and the contact impulse lies along the line
of centres. Contact occurs at frame $20.03 \pm 0.02$ in every clip; pre- and post-contact
velocities are read at frames $18$ and $22$, far enough from contact that the solver's
interpenetration window is excluded and close enough that no other event intervenes. We
write $v_A, v_B$ for the pre-contact velocities, $w_A, w_B$ for the post-contact ones,
$\Delta v_A = w_A - v_A$, $m_A, m_B$ for the masses and $n$ for the unit contact normal.

\paragraph{Per-object attributes.}
These belong to one body and are probed from that body's own token pool
(Appendix~\ref{sec:single}).

\begin{itemize}\itemsep2pt
\item \textbf{Velocity} $v_A \in \mathbb{R}^2$, the pre-contact velocity vector.
\item \textbf{Speed} $|v_A|$, its Euclidean norm. Note this is a \emph{norm} of the
      previous entry, not a linear function of it; the distinction matters
      (Appendix~\ref{sec:single}).
\item \textbf{Direction} $v_A / |v_A|$, the unit heading. On these clips it is $0.87$
      linear in $v_A$, since the speed distribution is narrow.
\item \textbf{Radius} $r_A$, the sphere's radius in world units, sampled independently per
      object.
\item \textbf{Mass} $m_A = \rho_A \tfrac{4}{3}\pi r_A^3$, probed as $\log m_A$. Density
      $\rho_A$ is drawn independently of radius and is signalled by the rendered material,
      so mass is visible but only through a cubed length and an appearance cue.
\end{itemize}

\paragraph{Between-object quantities.}
These physical quantities are central to what we examine. 

\begin{itemize}\itemsep2pt
\item \textbf{Momentum transfer}
      $\Delta p = m_A \Delta v_A + m_B \Delta v_B \in \mathbb{R}^2$. Zero exactly when
      momentum is conserved. It is a \emph{mass-weighted sum} over the two objects, which
      is what makes it legible frame-wide and absent object-locally
      (\S\ref{sec:momentum}).
\item \textbf{Momentum conservation}, the binary $|\Delta p| > 0$. A threshold on a norm:
      conserving clips sit at the origin and violating clips on a ring around it, so no
      linear map of $\Delta p$ separates them (\S\ref{sec:momentum}).
\item \textbf{Restitution}
      $e = |(w_A - w_B)\!\cdot n| \,/\, |(v_A - v_B)\!\cdot n|$, the ratio of separation to
      approach speed along the contact normal. \emph{Mass-free}: two visible speeds and
      nothing else. $e = 1$ is elastic, $e < 1$ inelastic, and $e > 1$ physically
      impossible.
\item \textbf{Elasticity}, the binary $e < 1$ (\S\ref{sec:elasticity-full}).
\item \textbf{Kinetic energy} $KE = \tfrac12 m_A |v_A|^2 + \tfrac12 m_B |v_B|^2$, and the
      ratio $KE_{\mathrm{post}}/KE_{\mathrm{pre}}$. Unlike restitution it cannot be formed
      without mass.
\item \textbf{Energy creation}, the binary $\Delta KE > 0$. One-sided and linear in the
      signed $\Delta KE$, the mirror of momentum conservation, which is why the two need
      opposite readouts (\S\ref{sec:momentum}, \S\ref{sec:energy}).
\item \textbf{Contact normal} $n$, the unit vector along the line of centres at contact.
      Frictionless contact puts the impulse along it, so we derive it from $\Delta v_A$
      rather than from the analytic geometry; the two agree to $10^{-14}$.
\end{itemize}

\paragraph{Friction quantities.}
Probed on the incline and slide sets (Section ~\ref{sec:friction}), which contain one cube
and a floor rather than two spheres.

\begin{itemize}\itemsep2pt
\item \textbf{Friction coefficient} $\mu$, the Coulomb coefficient between cube and floor,
      sampled independently of every other factor. It belongs to the \emph{pair}, not to
      either surface.
\item \textbf{Incline angle} $\theta$, the slope of the plane, sampled independently of
      $\mu$. Visible as static image geometry.
\item \textbf{Down-slope acceleration} $a$, measured from the trajectory. On the incline
      $a = g(\sin\theta - \mu\cos\theta)$ with $g = 0.45$; on the flat floor
      $a = \mu g$ with $g = 1.0$, so there $\mu$ and $a$ are the same number up to a
      constant.
\item \textbf{Cube size}, the control: it faces the same domain gap as $\mu$ under transfer
      but is a visible static property, so it establishes that a transfer failure is
      informative rather than a broken comparison.
\end{itemize}

\section{Samples from Datasets}\label{sec:samples_dataset}

\begin{figure}[ht]
\centering
\includegraphics[width=\linewidth]{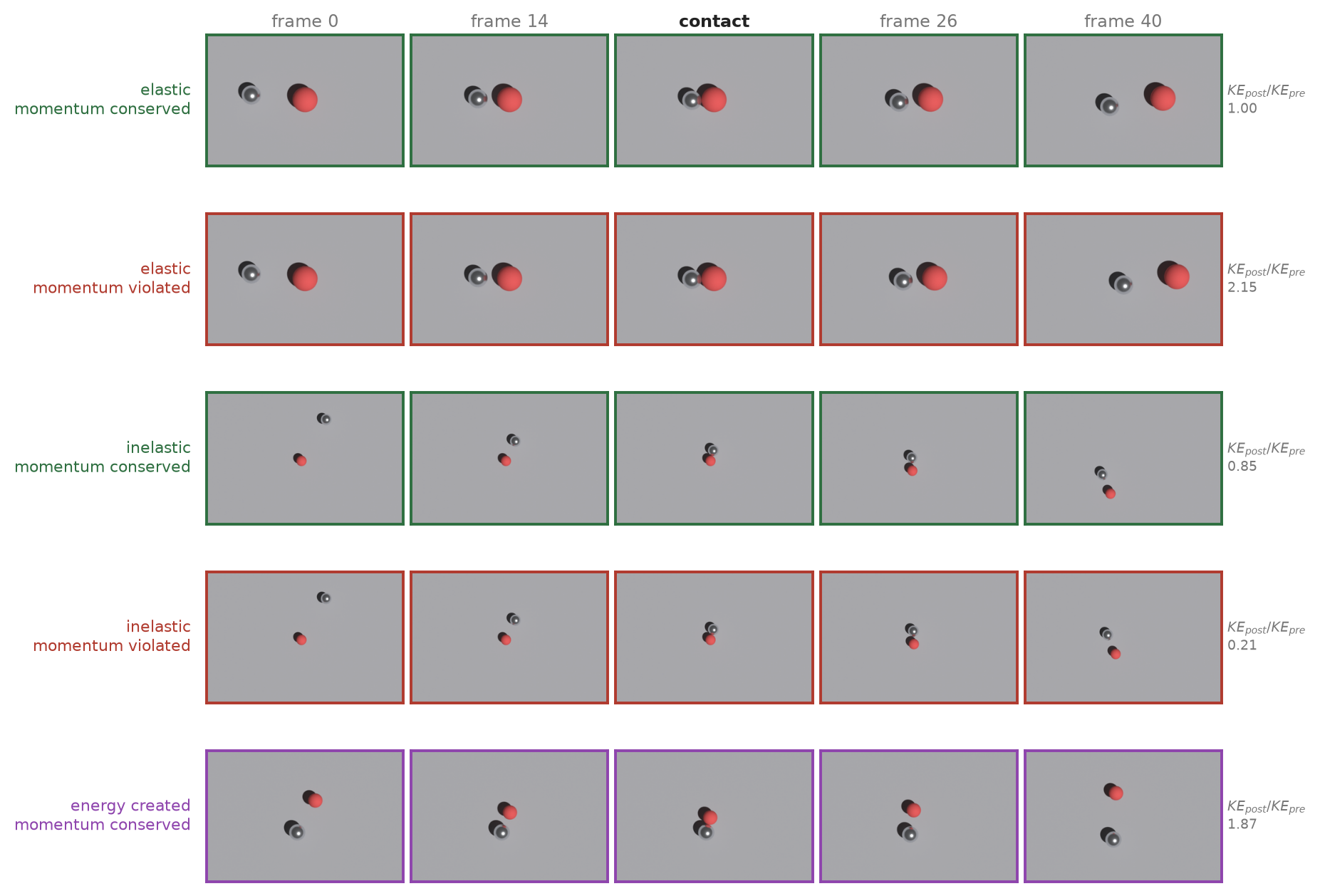}
\caption{\textbf{The five two-body regimes.} One clip per regime, cropped to the action;
contact falls at frame $20$ in every scene. Rows one and two are the \emph{same seed} with
and without the momentum edit, and share a crop, as do rows three and four: the members of a
pair are identical up to the contact frame and diverge only afterwards, so nothing before
contact distinguishes them. Green borders mark momentum-conserving clips, red
momentum-violating ones, purple the physically impossible arm. The ratio at the right of
each row is $KE_{\mathrm{post}}/KE_{\mathrm{pre}}$. Reading that column shows why the energy
experiment needs its own stimuli: the momentum edit moves kinetic energy as a side effect
(the elastic violating twin lands at $2.15$), so energy creation cannot be attributed to
anything on the momentum axis. Only the last row creates energy while conserving momentum
exactly.}
\label{fig:dataset-grid}
\end{figure}


\end{document}